\documentclass{article}
\usepackage{spconf,amsmath,graphicx}
\usepackage{graphicx,floatrow,subfig}
\usepackage{amsmath,textcomp,amssymb}
\usepackage[ruled]{algorithm2e}
\usepackage{slashbox}
\graphicspath{{BAMS/}}
\usepackage{array}
\label{1mame and address}
\title{A Bidirectional Adaptive Bandwidth Mean Shift Strategy For Clustering}
\name{Fanyang Meng $^{\star}$ \quad Hong Liu$^{\star}$ \quad  Yongsheng Liang$^{\dagger}$ \quad Liu Wei$^{\dagger}$ \quad Jihong Pei$^{\S}$  
\thanks{This work is supported by National High Level Talent Special Support Program, National Natural Science Foundation of China (NSFC,No.61340046,61673030,61331021,U1613209,U1301251), Shenzhen Science and Technology Project (JCYJ20160608151239996).}}
\address{$^{\star}$ Machine Perception,	Shenzhen Graduate School, Peking University, China \\
	$^{\dagger}$ ShenZhen Institute of Information Technology, China\\
	$^{\S}$ College of Information Engineering, Shenzhen University, China\\
 Email: fymeng,hongliu@pkusz.edu.cn; liangys,liuwei@sziit.edu.cn; jhpei@szu.edu.cn}
\begin{document}
\maketitle
\begin{abstract}
The bandwidth of a kernel function is a crucial parameter in the mean shift algorithm. This paper proposes a novel adaptive bandwidth strategy which contains three main contributions. (1) The differences among different adaptive bandwidth are analyzed. (2) A new mean shift vector based on bidirectional adaptive bandwidth is defined, which combines the advantages of different adaptive bandwidth strategies. (3) A bidirectional adaptive bandwidth mean shift (BAMS) strategy is proposed to improve the ability to escape from the local maximum density. Compared with contemporary adaptive bandwidth mean shift strategies, experiments demonstrate the effectiveness of the proposed strategy.	
\end{abstract}
\section{Introduction}
Mean shift is a nonparametric mode seeking algorithm \cite{Carreira-Perpinan2015,ComaniciuMeer-6}, which iteratively locates the modes in the data by maximizing the kernel density estimate. As with its the nonparametric nature, the mean shift algorithm becomes a powerful tool to mode-seeking and clustering \cite{YuanHu-48,GeorgescuShimshoni-40}, and it has also been applied to solve several computer vision problems, e.g., image filtering\cite{Carreira-Perpinan2015},  segmentation \cite{LiuZhou-39,MayerGreenspan-33,Carreira-Perpinan-50}, visual tracking  \cite{Collins-20,MohammadiAmoozegar-30,DulaiStathaki-23,SunYao-17,ZhaoWang-18} and action recognition \cite{Shan2007,Liu2017}.

The bandwidth of a kernel function is a crucial parameter in the mean shift algorithm \cite{ComaniciuMeer-6,Comaniciu-32,MingCi-43,ComaniciuRamesh-31}. Because of the intrinsic limitations of the fixed bandwidth mean shift, many adaptive-bandwidth-mean-shift (AMS) algorithms have been proposed \cite{MingCi-43,MayerGreenspan-33}. From the definition of bandwidth, the AMS algorithms can be generalized into two main strategies \cite{ComaniciuRamesh-31}: bandwidth variable with estimate point (hereafter, called the EAMS strategy) and bandwidth variable with sample point (hereafter, called the SAMS strategy).

In the EAMS strategy \cite{Collins-20,NingZhang-8,YuTian-7,WangYan-45}, because the weight of each sample point reciprocal to distance between estimate point and sample point. Therefore, the EAMS strategy is still satisfied in this case with the neighborhood constraints,and has good convergence and stability as a fixed bandwidth mean shift. However, because the bandwidth is still identically scaled for kernels centred for all of the sample points, its performance improvement over the fixed bandwidth mean shift is insignificant in this case \cite{ComaniciuRamesh-31}.

In the SAMS strategy \cite{MayerGreenspan-33,Comaniciu-32,ComaniciuRamesh-31}, the most attractive property is that a particular bandwidth choice considerably reduces the bias while the variance remains theoretically unchanged. However, on one hand, the weight of each sample point will not satisfy in this case with the neighborhood constraints, the SAMS strategy has worse convergence and stability than  EAMS strategy. On the other hand, because a new weights with sample point density has been introduced, it causes the SAMS strategy to easily fall into the local maximum density.

The two AMS strategies can be considered to optimize the weights of the sample points by adaptively adjusting the bandwidth. However, the difference between EAMS strategy and SAMS strategy has not received sufficient attention. In fact, the weight of each sample point under different AMS strategies has been significantly different (as shown in Fig.~\ref{image1} and the further analysis will be given in section \textbf{2}).


In this paper, a mean shift based on bidirectional adaptive bandwidth (BAMS) is proposed. The main contributions of this paper are summarized as follows.
\begin{enumerate}
	\item 
The mean shift algorithms under different ASM strategies are analyzed. It is found that the weights of sample points under different ASM strategies have significant difference.
	\item
The BAMS strategy combining the advantages of different AMS strategies is proposed to improve the ability to escape from the local maximum destiny. 
	\item
Experimental comparisons on a synthetic dataset and a benchmark dataset, i.e. BSD500 image dataset \cite{Arbelaez2011}, verify the effectiveness of the proposed BAMS strategy.	
\end{enumerate}

The remainder of this paper is structured as follows: the mean shift algorithms under different ASM strategies are analyzed in Section \textbf{2}; the BAMS strategy is proposed in Section \textbf{3}; Section \textbf{4} presents the experimental results; Section \textbf{5} concludes the paper. 
\section{Related Work}
 \begin{figure*}[htb]
 \begin{minipage}[b]{1.0\linewidth}
   \centering
   \centerline{\includegraphics[width= 13 cm,height= 7.5 cm]{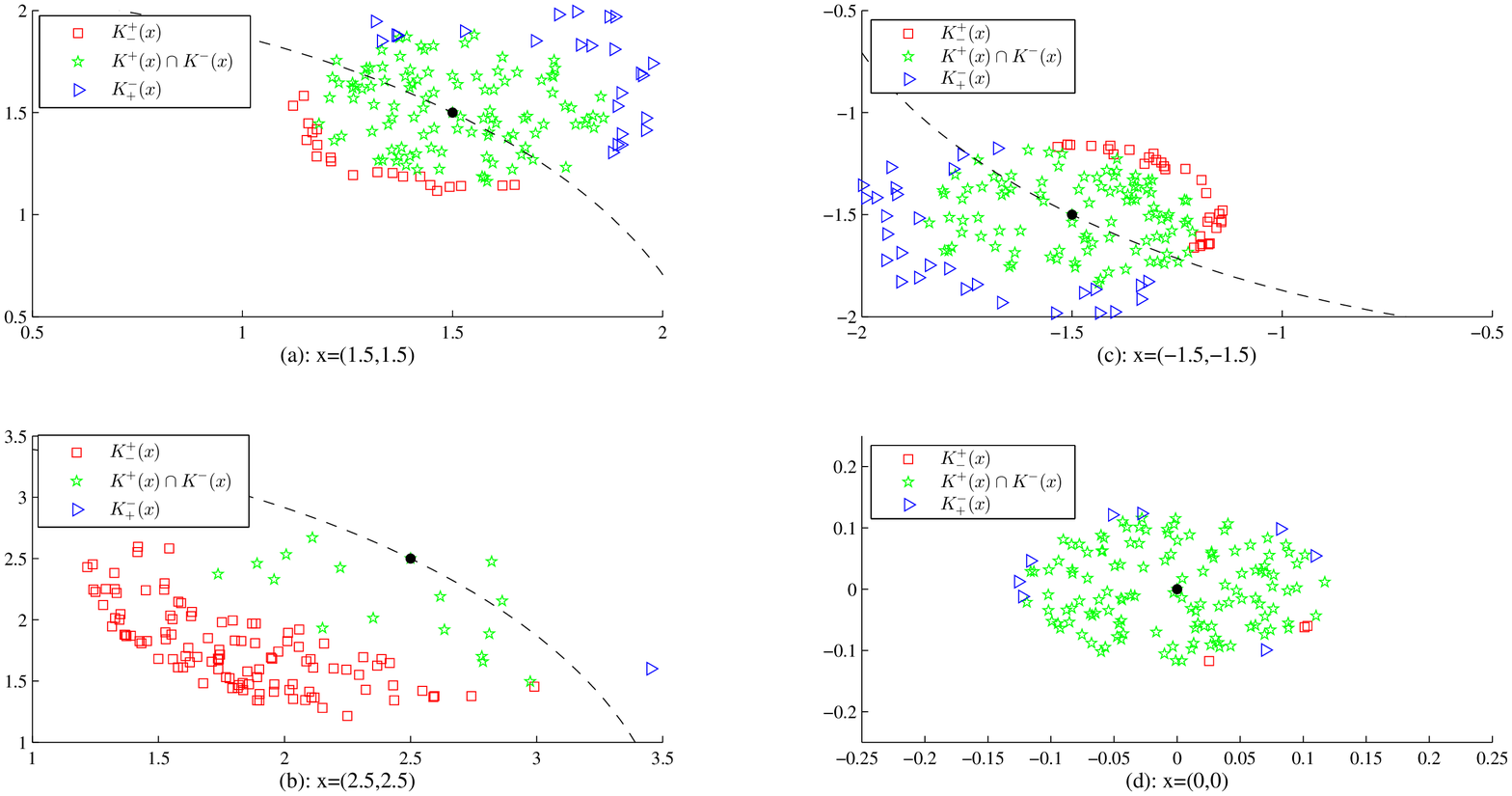}}
 \end{minipage}
  \caption{A simple illustration of the sample points distribution in the Positive \textit{knn} set $\textsl{K}^+$ and Negative \textit{knn} set $\textsl{K}^-$ in 2-d normal distribution. Here, $\textsl{K}^+_-(x) =  \textsl{K}^+(x) \cap  {\overline{\textsl{K}^-(x)}}$, $\textsl{K}^-_+(x) =  \textsl{K}^-(x) \cap  {\overline{\textsl{K}^+(x)}}$, and the black dotted lines is the contour line with $f(x)$.}
  \label{image1}
\end{figure*}
Given $n$ data points ${\mathop{X}\nolimits}= \{{{\mathop{x}\nolimits} _i}\}_{i=1,..n}$ and ${{\mathop{x}\nolimits} _i}$ with a bandwidth ${h_i}$ on a $d$-dimensional space, $i = 1,....n$, the sample point density estimator obtained with the kernel profile $k(x)$ is given by:
\begin{align}
\hat f({\mathop{x}\nolimits} ) = \frac{1}{n}\sum\limits_{i = 1}^n {\frac{k({\left\| {\frac{{{\mathop{x}\nolimits}  - {{\mathop{x}\nolimits} _i}}}{{{h_i}}}} \right\|^2})}{{h_i^d}}}
\label{NEq1}
\end{align}

Then, the mean shift vector is represented as follows:
\begin{align}
{{\mathop{\rm m}\nolimits} _{h_i}}(x) = \frac{{\sum\nolimits_{i = 1}^{n} {{h_i}^{-(d+2)}\cdot g({{\left\| {\frac{{{\mathop{x}\nolimits}  - {{\mathop{x}\nolimits} _i}}}{{{h_i}}}} \right\|}^2}){x_i}} }}{{\sum\nolimits_{i = 1}^n {{h_i}^{-(d+2)} \cdot g({{\left\| {\frac{{{\mathop{x}\nolimits}  - {{\mathop{x}\nolimits} _i}}}{{{h_i}}}} \right\|}^2})} }} - x
\label{NEq3}
\end{align}
where $g(x)= -k'(x)$.

Although there are numerous methods described in the statistical literature to define bandwidth and kernel function for the mean shift strategy, the simplest and most commonly way to obtain the bandwidth and kernel function are the k-nearest-neighbours (\textit{knn}) and the \textit{\textbf{Epanechnikov}} kernel functions \cite{MayerGreenspan-33}. Then, the AMS vectors with EAMS and SAMS strategies are rewritten as follows:
\begin{align}
{\mathop{\rm m}\nolimits} _{h_x}(x) &= \frac{1}{k}\sum\limits_{{x_i} \in {\textsl{K}^+}(x)} {{x_i}}  - x \label{EQ3} \\
{\mathop{\rm m}\nolimits} _{h_{x_i}}(x) &= \frac{\sum\limits_{{x_i} \in {\textsl{K}^-}(x)} {w_i \cdot x_i} }{{\sum\limits_{{x_i} \in {\textsl{K}^-}(x)} {w_i}}} - x
\label{EQ4}
\end{align}
where $w_i = h_{x_i}^{-(d+2)} $, ${\textsl{K}^+}({\bf{x}})$ is the Out \textit{knn} set of ${\bf{x}}$ ( hereafter, called the out \textit{knn} with $x$) and ${\textsl{K}^-}({\bf{x}})$ is satisfied with ${\textsl{K}^-}({\bf{x}}) = \{ {{\bf{x}}_j}|\left\| {{{\bf{x}}_j} - {\bf{x}}} \right\| \le {h_j}\}$ ( hereafter, called in \textit{knn} with $x$) \cite{FanyangMeng2015}. From Eq.~\ref{EQ3} and \ref{EQ4}, it can easily be seen that under different AMS strategies, the iut \textit{knn} set and in \textit{knn} set play an important role for the AMS vector.

Fig.~\ref{image1} gives a simple example to illustrate the difference between the out \textit{knn} set and in \textit{knn} set under different estimate points. It can be seen that the sample points in $\textsl{K}^+(x)$ and $\textsl{K}^-(x)$ have significant difference. Most of the sample points in $\textsl{K}^+(x)$ have larger probability density than the sample points in $\textsl{K}^-(x)$. It also can explain that why the SAMS strategy is vulnerable to be influenced by noises and can easily fall into the local maximum density in practice.
\section{Bidirectional Adaptive Bandwidth Mean Shift Strategy}
In this section, a new bidirectional AMS vector is defined to distinguish the existing AMS vector ( The previous AMS vector is called the unidirectal AMS vector). Then, a novel bidirectional AMS strategy is proposed, and the detailed implementation of bidirectional AMS strategy is provided.
\begin{algorithm}[t]
	\caption{The Pseudo Code of MS based on BAMS Strategy}
	\KwIn{$X$,$k$,$\lambda$,$\beta$}
	\KwOut{$X$}
	\Repeat{stop}
	{
		\For{$i= 1:n$}
		{
			\begin{flalign}
			&\forall i : d_j = g(\left\| \frac{x_i-x_j}{h_{x_j}}\right\|^2) - \lambda \cdot g(\left\|\frac{x_i-x_j}{h_{x_i}}\right\|^2) & \nonumber
			\end{flalign}
			$w_j = |d_j|/S(\beta \cdot d_n)$ \\
			$y_i \leftarrow \frac{\sum\limits_{j=1}^n w_j \cdot x_j}{\sum\limits_{i=1}^n w_j}$		
		}
		$\forall x_i \leftarrow y_i$
	}
	\medskip
	\textbf{{ Matrix Form}} \\
	\medskip
	\Repeat{stop}
	{	
		\begin{flalign}
		&G = (g(\left\| \frac{x_j-x_i}{h_{x_i}}\right\|^2))_{ij}& \nonumber
		\end{flalign}	
		$D = G - \lambda G^T$\\
		$W = |D|\mathbf{.}/S(\beta \cdot D)$\\
		$N = diag(\sum\limits_{i=1}^n{w_{ij}}) $\\
		$P = W/N$\\
		$X = XP$
	}
\end{algorithm}
\subsection{Bidirectional Adaptive Bandwidth Mean Shift Vector}
As shown in Fig.~\ref{image1} (a), (b) and (c), the densities of sample points in a set $ \textsl{K}^+_-(x)$ (red squares as shown in Fig~.\ref{image1}) are greater than $f(x)$. Therefore, the vector from $x$ to the centroid of set $\textsl{K}^+_-(x)$ (hereafter, called  positive mean shift vector) is points toward the direction of the density increase. By comparison, the densities of sample points in a set  $ \textsl{K}^-_+(x)$ (bule triangles as shown in Fig~.\ref{image1}) are lower than $f(x)$. It means that the vector from $x$ to the centroid of set $\textsl{K}^-_+(x)$ (hereafter, called negative mean shift vector) is points toward the direction of the density decrease.

From the above analysis, a bidirectional AMS vector of ${{\bf{x}}}$ is defined as follows:
\begin{align}\label{eq5}
	&M_{h_{x,x_i}}(x)  = \frac{\sum\limits_{i=1}^n{w_i\cdot x_i}}{\sum\limits_{i=1}^n{w_i}} - x \\
	&w_i  =   g(\left\| \frac{x-x_i}{h_x}\right\|^2) - g(\left\| \frac{x-x_i}{h_{x_i}}\right\|^2)
\end{align}

Fig.~\ref{image1} gives a simple example to further illustrate
the bidirectional AMS vector. Here, the bandwidth and kernel function are obtained by the \textit{knn} distance and the \textit{\textbf{Epanechnikov}} kernel functions. In this case, $w_i$ in Eq.~\ref{eq5} can be rewrited as follows:
\begin{align}\label{eq7}
w_i & =
\left\{
\begin{aligned}
	1 &  & if x_i \in \textsl{K}^+_-(x) \\
	-1 & & if x_i \in \textsl{K}^-_+(x)   \\
	0 &   & else
\end{aligned} \right.
\end{align}

Eq.~\ref{eq7} shows that there are two parts in the weight of each sample pint: the positive weight and the negative weight. on one hand, some sample points with higher density are considered in the positive weights. This ensures that the bidirectional mean shift vector points toward the direction of density increase. On the other hand, the negative weights take more information of the sample points with lower density into account. It ensures that the bidirectional mean shift vector is far from the direction of density decrease. Therefore, the bidirectional mean shift vector can fully utilize the information of the sample points and has more ability to escape from the local maximum density.
\subsection{Bidirectional Adaptive Bandwidth Mean Shift Strategy}
From the definition of the bidirectional mean shift vector, a new Mean Shift strategy (hereafter, called BAMS strategy) is designed. The BAMS strategy is performed in an iterative manner. The following steps describe the BAMS strategy in detail.
\subsubsection{Weighted Bidirectional Adaptive Bandwidth Mean Shift Vector}
After the bandwidths with estimate point and sample points are obtained, the weight of each sample points can be computed from Eq.~6. Eq.~6 is reasonable in theory, but some problems will be encountered in practice.

First, when an estimate pint is the local maximal (as shown in Fig.~\ref{image1} d) or local minimal (for example, image smoothing area), most weights computed from Eq.~6 are close to zero in this case. It will cause the instability of the BAMS strategy.

Second, with the dimensional increase, the distribution of the sample point set will became sparser, and the direction of negative mean shift vector has greater uncertainty. Additionaly, in an extreme case, the weights for all sample points are not zero, while the sum of all the weights for all sample points near zero. Those will cause the instability and divergence of the BAMS strategy.

To avoid above problems, the weight coefficients and Sigmoid function are utilized for mapping the weights of smaple points to [0,1] and keeping the monotonicity of weights. Then, the final weight calculation formula in this paper is as follows:
\begin{align}
w_i  &= \frac{|d_i|}{1+e^{-\beta \cdot d_i}} \\
d_i  &=   g(\left\| \frac{x-x_i}{h_x}\right\|^2) - \lambda \cdot g(\left\| \frac{x-x_i}{h_{x_i}}\right\|^2) \nonumber
\end{align}
\subsubsection{Implementation}
For a given $d$-dimensional sample points set $X = \{x_i\} _{i = 1,..n}$ and an initialize estimate point $x_i$, the iterative procedure for mode detection based on the BAMS strategy is shown in \textbf{Algorithm 1}. It is worthing noting that the positive weight matrix only needs to be calculated by EAMS strategy (or SAMS strategy), and the negative weight matrix can be obtained by the transpose of positive weight matrix. 
\begin{figure}[tb]
	\begin{minipage}[b]{1.0\linewidth}
		\centering
		\centerline{\includegraphics[width=8.5cm,height=5cm]{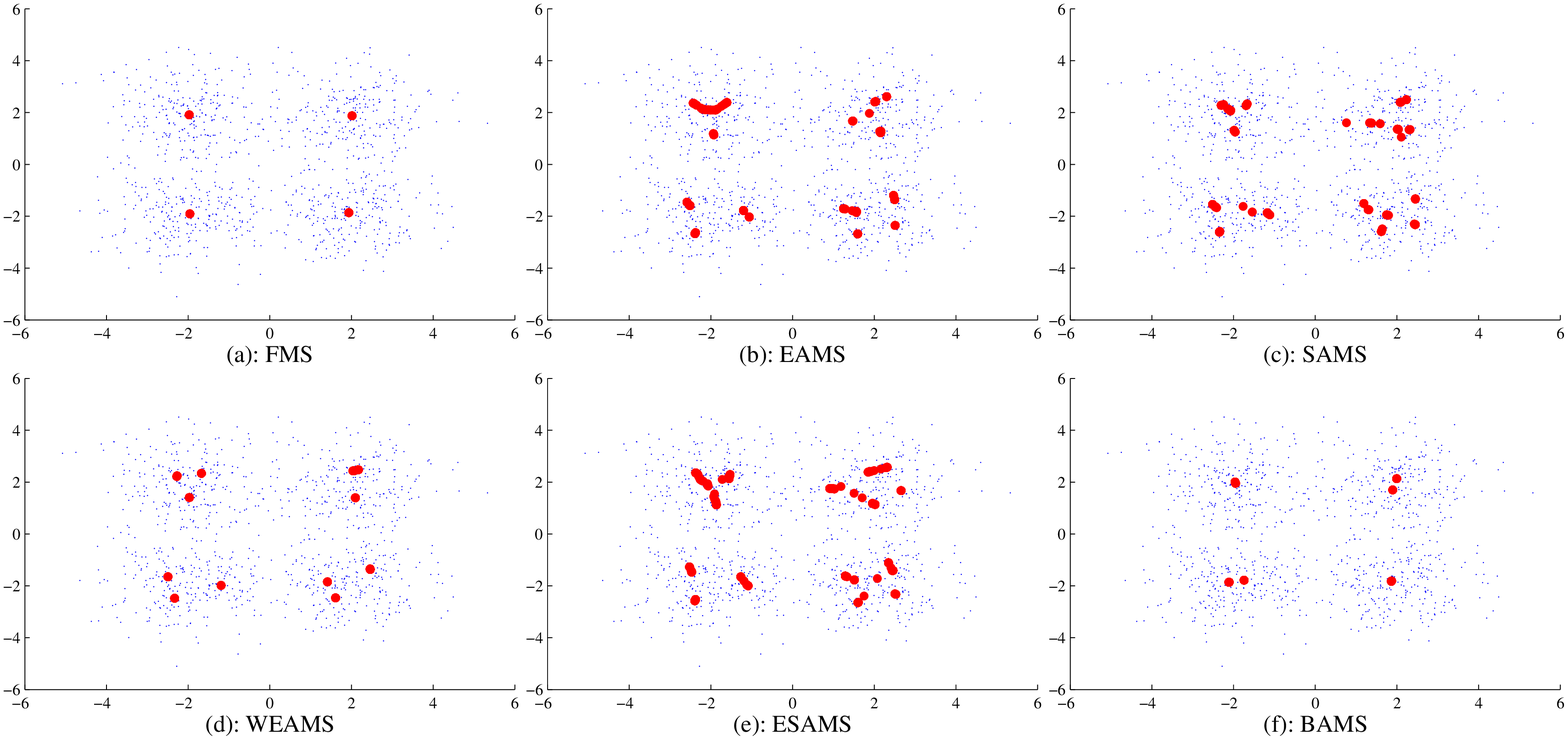}}
	\end{minipage}
	\caption{Performance comparisons of different AMS strategies with multiple Gauss distribution}
	\label{image2}
\end{figure}
\section{Experiments and Discussions}
To investigate the performance of the BAMS strategy, the results are compared to contemporary AMS strategies: \textbf{FMS} (Fixed bandwidth Mean shift strategy as a baseline), \textbf{EAMS}, \textbf{SAMS}, \textbf{ESAMS}, \textbf{WEAMS} and \textbf{BAMS}. Here, the weights in \textbf{ESAMS} strategy and \textbf{WEAMS} strategy are defined as follows:
\begin{align}
ESAMS &: g(\left\|\frac{x-x_i}{h_x}\right\|^2) + g(\left\| \frac{x-x_i}{h_{x_i}}\right\|^2) \nonumber \\
WEAMS &: {{h_{x_i}}^{-{d-2}}} \cdot g(\left\| \frac{x-x_i}{h_x}\right\|^2) \nonumber
\end{align}

In this paper, we performed experiments on two groups. The first group shows the results for the synthetic data, which is generated as 4-gaussian distribution in 2-d space. The second group represents the performance of all strategies on the BSD500 dataset \footnote{https://www2.eecs.berkeley.edu/Research/Projects/CS/vision/grouping \\ /resources.html}.

For the second group, the performance of different strategies is demonstrated on mean shift-based image filtering, The colour-level images in the L*u*v space were processed. To simplify, the gaussian kernel and the adaptive bandwidth with \textit{knn} distance are utilized for all experiments. In the proposed BAMS strategy, we just choose $\lambda$ and $\beta$ in Eq.~8 to be 0.975 and 100 for all experiments.

The first group results are depicted in Fig.~\ref{image2}. As shown in the Fig.~\ref{image2}, it can be seen that the proposed BAMS strategy is better than the EAMS other AMS strategies. Thus, the BAMS can effectively improve the ability of mean shift algorithm to escape from the local maximum density. It is worth noting that the ESAMS is worse than EAMS and SAMS, while the WEAMS is better than EAMS and SAMS. It is further explains the difference between EAMS and SAMS.
\renewcommand\arraystretch{1.25}	
\begin{figure}
	{
		\ttabbox{\caption{Evaluation of mean shift strategies on the BSDS300.BQ: Boundary quality; RQ: Region quality; PRI: Probability Rand Indx}}{
			\label{table}
			\centering
			\begin{tabular}
			{{cp{0.5cm}cp{0.5cm}cp{0.5cm}cp{0.5cm}cp{0.5cm}cp{0.5cm}c}}\hline		
			\multicolumn{1}{c}{Strategy}
			&\multicolumn{2}{c}{BQ}
			&\multicolumn{2}{c}{RQ}
			&\multicolumn{2}{c}{PRI} \\
		         & ODS  & OIS  & ODS  & OIS   & ODS  & OIS\\  \hline
		    FMS  & 0.57 & 0.61 & 0.52 & 0.59  & 0.80 & 0.83\\	\hline     
			EAMS & 0.60 & 0.62 & 0.59 & 0.63  & 0.84 & 0.86\\	\hline	
			SAMS & 0.61 & 0.64 & 0.57 & 0.60  & 0.83 & 0.85\\ \hline
			ESAMS& 0.60 & 0.61 & 0.55 & 0.58  & 0.83 & 0.85\\ \hline
			WEAMS& 0.56 & 0.59 & 0.54 & 0.55  & 0.82 & 0.83\\ \hline
			BAMS & 0.61 & 0.64 & 0.58 & 0.62  & 0.84 & 0.85\\ \hline
	\end{tabular}}}
\end{figure}

The second group results are depicted in Table.~\ref{table}. As shown in Table~\ref{table}, it can be observed that BAMS and SAMS are better than other strategies in term of boundary quality. Therefore, the BAMS and the SAMS can preserve more the edge information than the others strategies. Although the BAMS is worse than the EAMS in term of region quality, it is still better than the others strategies. It means that the BAMS is a good compromise in regional consistency property and boundary to property.  
\section{Conclusion and Future Work}
The bandwidth is a crucial parameter in the mean shift algorithm, which directly affects the performance. To give a better bandwidth selection strategy, the differences between EAMS and SAMS are carefully analyzed firstly. Then, a bidirectional adaptive bandwidth mean shift algorithm (BAMS) is proposed, which combines the advantages of the EMAS and the SMAS. Experiments show that the results obtained agree well with the theory. The theoretical analysis and optimization, e.g. extended to directed graph model, can be further improved in future work.
\bibliographystyle{ieeetr}
\bibliography{BAMSRef}
\end{document}